%% file: main.tex
\documentclass{article}
\usepackage{iclr21}

\usepackage{balance} 

\input{math_commands.tex}

\usepackage{times}
\usepackage{soul}
\usepackage{url}

\definecolor{blue}{rgb}{0.53, 0.81, 0.92}

\usepackage[colorlinks = true,
            linkcolor = red,
            urlcolor  = red,
            citecolor = blue,
            anchorcolor = blue]{hyperref}

\usepackage[utf8]{inputenc}
\usepackage[small]{caption}
\usepackage{graphicx}
\usepackage{amsmath}
\usepackage{amsthm}
\usepackage{booktabs}
\usepackage{algorithm}
\usepackage{algorithmic}
\usepackage{numprint}
\usepackage[american]{babel}
\usepackage{graphicx}
\usepackage{url}
\usepackage{xspace}
\usepackage{booktabs}       
\usepackage{amsmath,amssymb,amsfonts,dsfont}       
\usepackage{nicefrac}       
\usepackage{bm}
\usepackage{enumitem}
\usepackage{todonotes}
\usepackage{natbib}

\usepackage{wrapfig}
\usepackage{cleveref}
\Crefformat{section}{Section~#2#1#3}
\Crefformat{proposition}{Proposition~#2#1#3}
\Crefformat{equation}{Eq.\,#2#1#3} 

\let\oldcitep=\citep
\renewcommand{\citep}[1]{\textcolor[rgb]{0,0,1}{\oldcitep{#1}}}

\title{Don't Do What Doesn't Matter: Intrinsic Motivation with Action Usefulness}

\author{
	Mathieu Seurin$^1$, Florian Strub$^2$, Philippe Preux$^1$, Olivier Pietquin$^3$\\ 
	\\
	$^1$Univ. Lille, CNRS, Inria, Centrale Lille, UMR 9189 CRIStAL, F-59000 Lille, France\\
	$^2$ DeepMind, Paris, France\\
	$^3$ Google Research, Brain Team, Paris, France\\
	contact : \texttt{mathieu.seurin@inria.fr}
}

\definecolor{Ao}{rgb}{0.0, 0.5, 0.0}
\definecolor{ride}{RGB}{222.0, 143.0, 5.0}
\definecolor{count}{RGB}{213.0, 94.0, 0.0}
\definecolor{rnd}{RGB}{2, 158, 115}
\definecolor{bright}{RGB}{238, 129, 66}
\definecolor{dark}{RGB}{34, 33, 41}
\definecolor{green}{RGB}{0, 102, 51}
\definecolor{blue}{RGB}{0,0,255}
\definecolor{ram}{RGB}{1, 115, 178}

\newcommand{\ram}{DoWhaM\xspace}
\newcommand{\ramlong}{Don't Do What Doesn't Matter\xspace}

\usepackage{amsmath}
\usepackage{float}

\iclrfinalcopy 
\begin{document}

\maketitle

\begin{abstract}
Sparse rewards are double-edged training signals in reinforcement learning: 
easy to design but hard to optimize. Intrinsic motivation guidances have thus been developed toward alleviating the resulting exploration problem. They usually incentivize agents to look for new states through novelty signals. Yet, such methods encourage exhaustive exploration of the state space rather than focusing on the environment's salient interaction opportunities. We propose a new exploration method, called \ramlong (\ram), shifting the emphasis from state novelty to state with relevant actions. While most actions consistently change the state when used, \textit{e.g.} moving the agent, some actions are only effective in specific states, \textit{e.g.}, \emph{opening} a door, \emph{grabbing} an object. \ram detects and rewards actions that seldom affect the environment.
We evaluate \ram on the procedurally-generated environment MiniGrid, against state-of-the-art methods and show that \ram greatly reduces sample complexity.
\footnote{Source code available at \url{https://github.com/Mathieu-Seurin/impact-driven-exploration}}

\end{abstract}


\section{Introduction}

We consider the reinforcement learning (RL) problem in which an agent  learns to interact with its environment optimally w.r.t.\@ a cumulative function of reward signals collected along its trajectories~\citep{sutton2018reinforcement, kaelbling1996reinforcement}.
To do so, an RL agent explores its surrounding, aiming at retrieving the most prominent course of actions, and updates its behavior accordingly.
When the environment provides abundant rewards, the agent may successfully collect enough training signals by performing random actions~\citep{mnih2016asynchronous,lillicrap2015continuous}.
But as soon as the environment provides scarce rewards, the agent is reduced to inefficiently waver around without being able to update its policy. To palliate this lack of training signals, one common method consists in intrinsically motivating the agent to explore its environment using a self-rewarding mechanism~\citep{schmidhuber1991possibility,oudeyer2007intrinsic, oudeyer2008can, schmidhuber2010formal}.

In the online RL literature, a widespread strategy is to augment the sparse \emph{extrinsic} reward from the environment with a generated dense \emph{intrinsic} reward that steers exploration~\citep{chentanez2005intrinsically, csimcsek2006intrinsic}.
Hence, the intrinsic reward should encode a degree of “novelty,” “surprise,” or “curiosity”~\citep{berlyne1965structure}
which is often encoded as an estimate of the agent's visitation frequency of state-action pairs. The agent is  incentivized to diversely interact with its environment to collect intrinsic rewards, which may ultimately trigger extrinsic rewards. Nonetheless, establishing intrinsic motivation signals remains double-edged as it introduces human-priors~\citep{sutton2019bitter}, may lead to sub-optimal policies~\citep{amodei2016concrete}, or foster reward hacking behavior~\citep{ng1999policy}.

All in all, different novelty measures have been studied, where each of them entails different exploration behaviors. For instance, count-based methods keep counts of previous observations to bait the agent to explore unseen states~\citep{lopes2012exploration, ostrovski2017count, bellemare2016unifying,ecoffet2019go}. Yet, these approaches implicitly encourage an exhaustive search of the state space.

Differently, curiosity-based methods train a model that encapsulates the environment dynamics, before nudging the agent to visit state-transitions with high prediction errors~\citep{pathak2017curiosity,burda2018exploration,haber2018learning,houthooft2016vime} or large change in the value of state features~\citep{raileanu2019ride}. 
However, the first category suffers from reward decay across episodes and poor generalization within procedurally-generated environments. We here observe that the second category insufficiently favors exploration towards novel and useful actions. 

In this paper, we therefore aim to shift the emphasis from state novelty distributions towards novel action distributions to develop new intrinsic motivation signals, and consequently, change the exploration behavior. More precisely, we aim at encouraging the agent to visit states that allow rare and relevant actions, i.e. actions that can only be performed in rare occasions. 
Imagine that an infant discovers that pushing a button triggers a light; s/he is likely to push everywhere to switch on new lights. By repeating his/her action, the infant may eventually uncover new buttons, and start associating the action \textit{push} to the relevant state features of \textit{buttons}. A similar observation can be made within virtual environments and embodied agents. We expect the agent to first detect rare actions to learn while being nudged towards the states that allow performing such actions.

In this spirit, we propose a new approach we name \ramlong (\ram). Instead of uniformly seeking for novel states, \ram encourages exploring states allowing actions that are rarely useful; those rarely relevant actions are generally hard to retrieve by random exploration. In other words, the agent is intrinsically rewarded when successfully performing an action that is usually ineffectual. We observe that this simple mechanism induces a remarkably different exploration behavior differing from the common state-count and curiosity-based patterns. %

Formally, \ram keeps track of two quantities for each action: the number of times the action has been used and the number of times the action led to a state change. The resulting intrinsic reward is inversely proportional to the number of times the action has led to a state change. Noticeably, \ram primarily keeps count of actions, and can thus be defined as an action count-based method. Besides, tracking actions (as opposed to states) naturally scales in RL: in the discrete case, there is generally less than a few thousand actions, allowing for an exact count. In the continuous case, actions may easily be discretized without using complex density models~\citep{tang2020discretizing}. 
Furthermore, \ram also preserve curiosity-based flavors by rewarding surprising transitions without training a complex predictive model or facing reward fading issues~\citep{pathak2017curiosity}. 

This paper first provides an overview of recent exploration methods before introducing \ram as an action-driven intrinsic motivation method. We then study this approach in the MiniGrid procedurally generated environment~\citep{chevalier2018babyai}. Despite their apparent simplicity, these tasks contain intermediate decisive actions, e.g. picking keys, which have kept in check advanced exploration methods ~\citep{raileanu2019ride,campero2020learning}. We empirically show that \ram reduces the sample complexity by a factor of 2 to 10 in a diverse set of environments while resolving the hardest tasks. We then study how \ram amends the agent's behavior and compare it to other methods. Finally, we also analyze whether \ram may lead to unwanted agent behavior when facing environment with multiple interactions, which we refer as the \textit{BallPit-problem}.

\section{Related Works}

RL algorithms require the agents to acquire knowledge about their environment to update their policy; exploration has thus been one of the longest running problems of RL~\citep{sutton2018reinforcement, mozer1990discovering,sato1988learning,schmidhuber1991possibility,barto1991computational}.
Exploration methods have quickly been categorized into two broad categories: \emph{directed} and \emph{undirected} exploration~\citep{thrun1992efficient}.

On the one hand, undirected exploration does not use any domain knowledge and ensures exploration by introducing stochasticity in the agent's policy. This approach includes methods such as random walk, $\epsilon$-greedy, or Boltzmann exploration. Although they enable learning the optimal policy in the tabular setting, they require a number of steps that grows exponentially with the state space~\citep{whitehead1991complexity, kakade2003sample,strehl2006pac}. 
Despite this inherent lack of sample efficiency, they remain valuable task-agnostic exploration strategies in large-scale problems with dense rewards~\citep{mnih2016asynchronous}.

Besides, recent undirected exploration methods have been developed to fit deep neural architectures such as injecting random noise in the network parameter space~\citep{fortunato2018noisy,plappert2018parameter}. 
Yet, undirected methods still struggle with sparse reward signals~\citep{plappert2018parameter}, or any task requiring deep exploration~\citep{osband2016deep, kearns2002near}.

On the other hand, directed methods incorporate external priors to orient the exploration strategy through diverse heuristics or measures. Among others, uncertainty has been used to guide exploration towards ill-estimated state-action pairs by relying on the Bellman equation~\citep{geist2010kalman, o2018uncertainty} 
or by bootstrapping multiple Q-functions~\citep{osband2016deep}. Despite being theoretically sound, these methods face scaling difficulties. In this paper, we study another directed exploration approach based on reward bonuses to densify the reward signal. 

In this setting, the environment reward, namely \emph{extrinsic} rewards, is augmented with an exploration guidance reward signal, namely \emph{intrinsic} rewards~\citep{chentanez2005intrinsically, csimcsek2006intrinsic}. 
This intrinsic reward spurs exploration by tipping the agent to take a specific course of actions. Furthermore, it makes undirected exploration mechanism applicable again by spreading milestone rewards during training. Inspired by cognitive science, this intrinsic reward often encodes a degree of “novelty,” “surprise,” ,“curiosity” ~\citep{oudeyer2007intrinsic, berlyne1965structure,schmidhuber1991possibility} 
, “learning progress”~\citep{lopes2012exploration} or “boredom”~\citep{schmidhuber1991possibility, oudeyer2008can}. 
These common intrinsic motivation mechanisms are broadly categorized into three families: count-based, curiosity-based, and goal-based methods. 

Count-based exploration aims to catalog visited states (or action-states pairs) along episodes to detect unseen states, and drive the agent towards them. It has first been proposed as an exploration heuristic in the early days of RL~\citep{thrun1992efficient, sato1988learning,barto1991computational} before being framed as  an intrinsic exploration reward mechanisms in the tabular case~\citep{strehl2008analysis, kolter2009near}.
Pseudo-counts were then introduced to approximate the state counts~\citep{lopes2012exploration}, where pseudo-counts were estimated through different density models to produce intrinsic rewards. Density models range from raw image downscaling with or without handcrafted state features~\citep{ecoffet2019go}, contextual trees~\citep{bellemare2016unifying}, generative neural models, e.g. PixelCNN~\citep{ostrovski2017count}, or autoencoders combined with a local hashing function~\citep{tang2017exploration}. Differently, \citet{burda2018exploration} use the prediction error between a randomly initialized network and a trained network as a state-count proxy. The random network acts a pseudo-count proxy by modeling a locally preserving hashing function while the regression error diminishes with the visit count. \cite{zhang2020bebold} uses pseudo-counts to model a knowledge frontier and pushes the agent beyond.

Finally, other methods incorporate back the reward in the pseudo-count state representation by using value-state representations~\citep{martin2017count} or distance between two successor features~\citep{machado2018count}. 
Yet, count-based methods may explore the immediate surrounding and heavily depend on the state representation quality. By shifting the emphasis on counting action, we thus address these representation constraints and push for distant interactions.

Curiosity-based exploration aims to encourage the agent to uncover the environment dynamics rather than cataloging states. Inspired by cognitive science, such agents learn a world model predicting the consequences of their actions; and they take an interest in challenging and refining it~\citep{haber2018learning,oudeyer2007intrinsic, oudeyer2018computational}. 
In RL, this intuition is transposed by taking the current state and action to predict the next state representation; the resulting prediction error is then turned into the intrinsic reward signal. Approaches mostly differ in learning the state representation: \citet{stadie2015incentivizing} compress raw observation with autoencoders, \citet{burda2019largescale} use random projections,  \citet{houthooft2016vime} capture the environment stochasticity by maximizing mutual information with Bayesian Networks. In parallel, \citet{pathak2017curiosity} argue that the state representation should mainly encode features altered by the agent. They thus introduce an inverse model that predicts the action given two consequent states as a training signal.

Finally, \citet{achiam2017surprise,azar2019world} computes the intrinsic reward across multiple timesteps predictions to better estimate information gain.
Yet, those intrinsic rewards based on prediction errors may attract the agent into irrelevant yet unpredictable transitions. Another drawback is reward evanescence: the intrinsic reward slowly vanishes as the model is getting better. \citet{schmidhuber1991possibility} originally proposed to measure the mean error evolution rather than immediate errors to account for the agent progress. Differently, \citet{raileanu2019ride} replace the error prediction by the difference between consecutive representation states, removing the need to compute a vanishing prediction error. In this paper, we also compare successive states in a similar spirit, but we use it to catalog actions and bias state visitation through a different exploration scheme.

Goal-based methods~\citep{colas2020intrinsically} provide identifiable and intermediate goals to reward the agent upon completion. Such approaches perform an explicit curriculum by slowly increasing the exploration depth through goal difficulties. They often build on top of the UVFA framework to condition the agent policy~\citep{schaul2015universal}. Goal-based methods may take several forms ranging from hindsight experience replay~\citep{andrychowicz2017hindsight}, adversarial goal-generation~\citep{forestier2017intrinsically,campero2020learning, sukhbaatar2018intrinsic,florensa2018automatic}
and hand-crafted goals~\citep{hermann2017grounded}. 
Yet, they may face to unstable training, complex goal definition~\citep{cideron2019self}, or require fully observable environment~\citep{campero2020learning}.

Other forms of intrinsic reward have been explored with empowerment~\citep{mohamed2015variational, gregor2016variational} or trajectory diversities~\citep{savinov2018episodic}, but they are facing scalability issues. \citet{hussenot2020show} also tried to retrieve intrinsic motivation signals from human trajectories through inverse reinforcement learning~\citep{russell1998learning}.

Other works tried to compose with multiple intrinsic bonus by getting both inter and intra episodic reward mechanisms~\citep{badia2019never} and/or different bootstrapping heads~\citep{beyer2019mulex,badia2020agent57}. 

Finally, intrinsic motivation have been explored in hierarchical reinforcement learning~\citep{barto2004intrinsically,kulkarni2016hierarchical, dilokthanakul2019feature, zhang2019scheduled,gregor2016variational}, but it goes beyond the scope of this paper.

\section{Reinforcement Learning Background}

\paragraph{Notation} The environment is modeled as a Markov Decision Problem (MDP), where the MDP is defined as a tuple $\{\mathcal{S},\mathcal{A},\mathcal{P},\mathcal{R},\gamma\}$.
At each time step $t$, the agent is in a state $s_{t} \in \mathcal{S}$, where it selects an action $a_t \in \mathcal{A}$ according to its policy $\pi:  \mathcal{S} \rightarrow \mathcal{A}$. It then receives a reward $r_{t}$ from the environment's reward function $r: \mathcal{S} \times \mathcal{A} \times \mathcal{S}  \rightarrow \mathds{R}$ and moves to the next state $s_{t+1}$ with probability $p(s_{t+1}|s_{t},a_{t})$ according to the transition kernel $\mathcal{P}$. 
Hence, the agent generates a trajectory $\tau=[s_0, a_0, r_0, s_1, r_1, a_1, \dots, s_T, a_T, r_T]$ of length $T$.
In practice, the policy is often parameterized by a weight vector $\bm{\theta} \in \Theta$. The goal is then to search for the optimal policy $\pi_{\bm{\theta}^*}$ that maximizes the expected return $J(\bm{\theta}) = E^{\pi_{\bm{\theta}}}\big[\sum_{t=0}\gamma^{t}r(s_t,a_t,s_{t+1}) \big]$ by directly optimizing the policy parameters $\bm{\theta}$. 

\paragraph{Intrinsic Motivation} In this setting, the reward function is decomposed into an extrinsic reward returned by the environment $r^e(s_t,a_t)$ and a new intrinsic reward $r^i(s_t,a_t,s_{t+1})$. Therefore, the new reward function is defined as $r(s_t, a_t,s_{t+1}) = r^{e}(s_t,a_t,s_{t+1}) + \beta r^{i}(s_t,a_t,s_{t+1})$ where $\beta$ is an hyperparameter to balance the two return signals. In practice, the extrinsic reward is often a sparse task-specific signal while the intrinsic reward is usually a dense training signal that fosters exploration. 

\begin{figure}[t!]
        \centering
    	\includegraphics[width=0.8\linewidth]{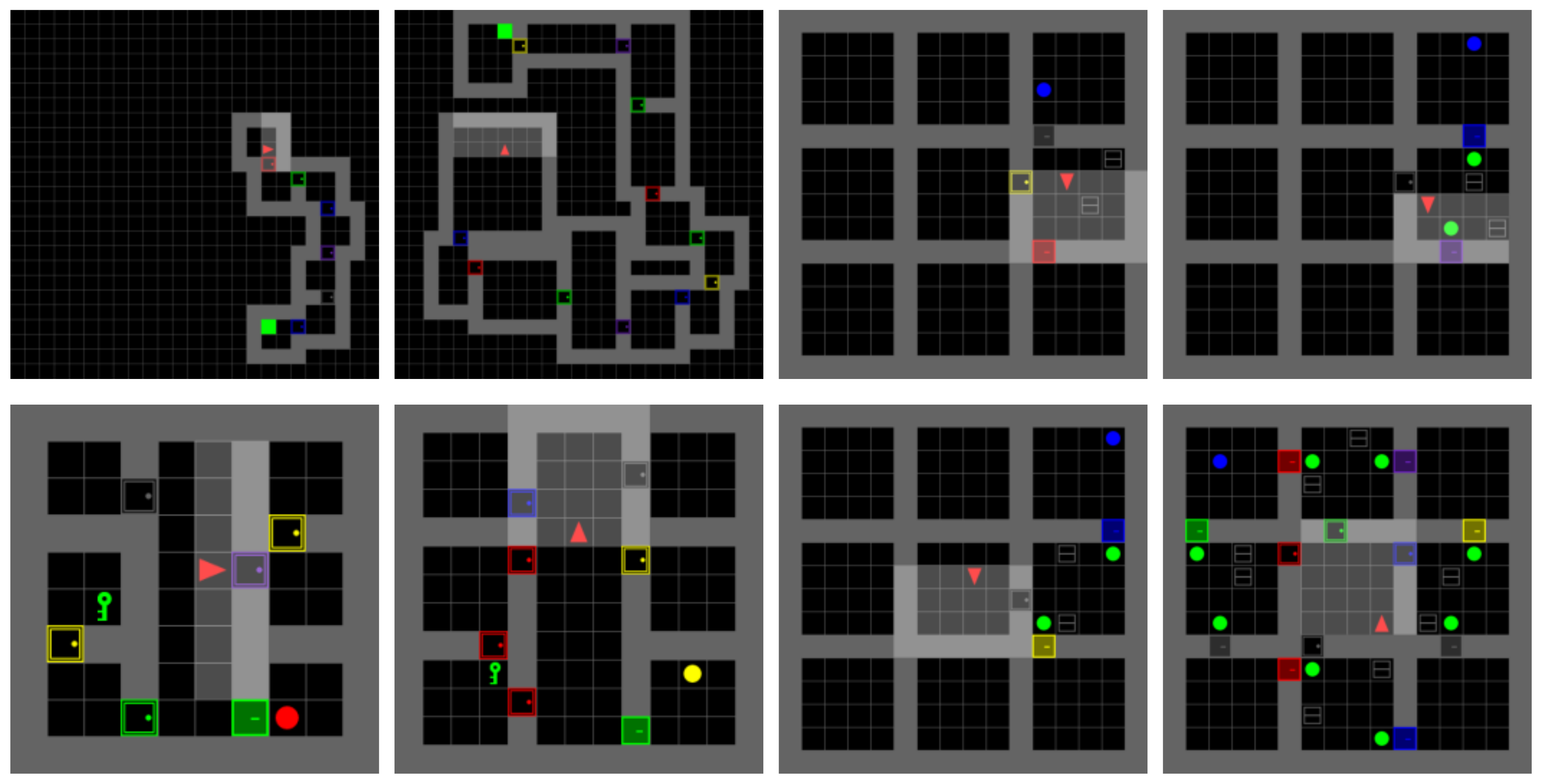}
    	\caption{Minigrid instances used, from top left to down right : MultiRoom (N7S4, N12S10), ObstructedMaze (2Dlh, 2Dlhb), KeyCorridor (S4R3, S5R3), ObstructedMaze (1Q, Full)}
    	\label{fig:envs_grid}
\end{figure}

\section{\ramlong}

\paragraph{Intuition}

While most actions consistently move the agent to a new state, some actions do not affect specific states, i.e., the agent remains in the same state after performing it. We hence define an \textit{effective action} if the new state of the environment is different from what it would have been if no action were to be taken. For instance, in tasks involving embodied interaction, such state-action pairs include  moving forward while facing a wall or grabbing non-existent objects. Although one may update the MDP only to keep effective actions, such an operation may not always be feasible or desirable in practice. It is thus up to the agent to learn the correct actionable states through exploration. Noticeably, those rare state-actions are often landmarks in the environment dynamics, e.g., triggering buttons or opening doors. One idea is thus to bias the agent to visit states that effectively allow rare actions. \ram encapsulates this exploration pattern by (1) detecting rare but effective actions, (2) rewarding the agent when effectively performing these rare actions. In short, rare and effective actions are the relevant actions that matter.



\paragraph{Method}

For every action $a_i$, the agent tracks two quantities. The number $U$ of times an action has been \textbf{u}sed during past trajectories, and the number $E$ of times the action was \textbf{e}ffective, i.e. change the state $s_t \neq s_{t+1}$. 
Formally, given the whole history of transitions (across all episodes) $\mathcal{H} = (s_h, a_h, s_{h+1})_{h=0}^H$: 
\begin{align}
 U^{\mathcal{H}}(a) &=  \sum_{h=0}^{H} \mathbf{1}_{\{a_h = a\}}, \\
 E^{\mathcal{H}}(a) &= \sum_{h=0}^{H} (\mathbf{1}_{\{a_h = a\}} \times \mathbf{1}_{\{s_h \neq s_{h+1}\}} ),
\end{align}
where $\mathbf{1}$ is the indicator function and $\times$ the product operator. 

Intuitively, the ratio $E^{\mathcal{H}}(a)/U^{\mathcal{H}}(a)$ indicates how often the action $a$ has been effective along the history $\mathcal{H}$. For instance, actions that move an agent would update the state most of the time, therefore $U(a_{i})\approx E(a_{i})$. On the other hand, grabbing objects only changes the state in rare occurrences, and $U(a_{i}) \ge E(a_{i})$. We then define the bonus as:


\begin{equation}
	B(a_t)= 
		\frac{
		    \eta^{1 - \frac{E^{\mathcal{H}}(a_t)}{U^{\mathcal{H}}(a_t)}} - 1
		    }
		    {\eta - 1},
\end{equation}
  
where $\eta$ is a hyperparameter. This function is a continuous approximation of an exponential decay $\exp^{-\eta E^{\mathcal{H}}(a_t)/U^{\mathcal{H}}(a_t)}$. It ranges from 1 when $E^{\mathcal{H}}=0$ and goes to 0 when $E^{\mathcal{H}}=U^{\mathcal{H}}$. Small $\eta$ leads to a uniform bonus on all actions whereas, large values favor rare and efficient actions (see \autoref{app:decay} in appendix for an illustration).



An intrinsic reward mechanism often requires to discount the intrinsic bonus within an episode. Hence, it prevents the agent from overexploiting, and being stuck in local exploration minima. Inspired by theoretically sound count-based methods~\citep{strehl2008analysis}, we thus divide the previous ratio by an episodic state-count.

Finally, we want to reward action only in context where they are effective, thus the agent is rewarded only when $st \neq s_{t+1}$, defining the final \ram intrinsic reward:
\begin{equation}
r^{i}_{\text{\ram}}(s_t, a_t, s_{t+1}) = 
\begin{cases}
	\frac{B(a_t)}{\sqrt{N^{\tau}(s_{t+1})}} & \text{if } s_t \neq s_{t+1} \\
	0 & \text{otherwise} \\
	\end{cases}
\end{equation}
where $N^{\mathcal{\tau}}(s) = \sum_{h=0}^{t} \mathbf{1}_{\{s = s_h\}}$ is an episodic state count which is reset at the beginning of each episode. In high-dimensional state space, the episodic state count can be replaced by a pseudo-count~\citep{bellemare2016unifying} or an episodic novelty mechanism~\citep{badia2020agent57}.

\textbf{Action-based Counter} As counting methods may sound anachronistic, we emphasize again that actions are ascertainable in RL, i.e. they can be easily enumerated. As opposed to state-counting which requires complex density models~\citep{ostrovski2017count}, discrete action suffers less from the curse of dimensionality, and can easily be binned together in the case of a large action set~\citep{dulac2015deep}. Besides, although \ram relies on an episodic state count, a raw approximation is sufficient as it encodes a reward decay. 

\section{Experimental settings}

We evaluate \ram in the procedurally-generated environments MiniGrid~\citep{chevalier2018babyai}.
MiniGrid is a partially observable 2D gridworld with a diverse set of tasks. The RL agent needs to collect items and open locked doors before reaching a final destination. Despite its apparent simplicity, several MiniGrid environments require the agent to perform exploration with few specific interactions, and have kept in check state-of-the-art exploration procedures~\citep{raileanu2019ride}. For each experiment, we report the rolling mean (over 40k timesteps) and standard deviation over 5 seeds. 


\subsection{MiniGrid Environment}\label{subsec:env}

\begin{figure*}
	\centering
	\includegraphics[width=\linewidth]{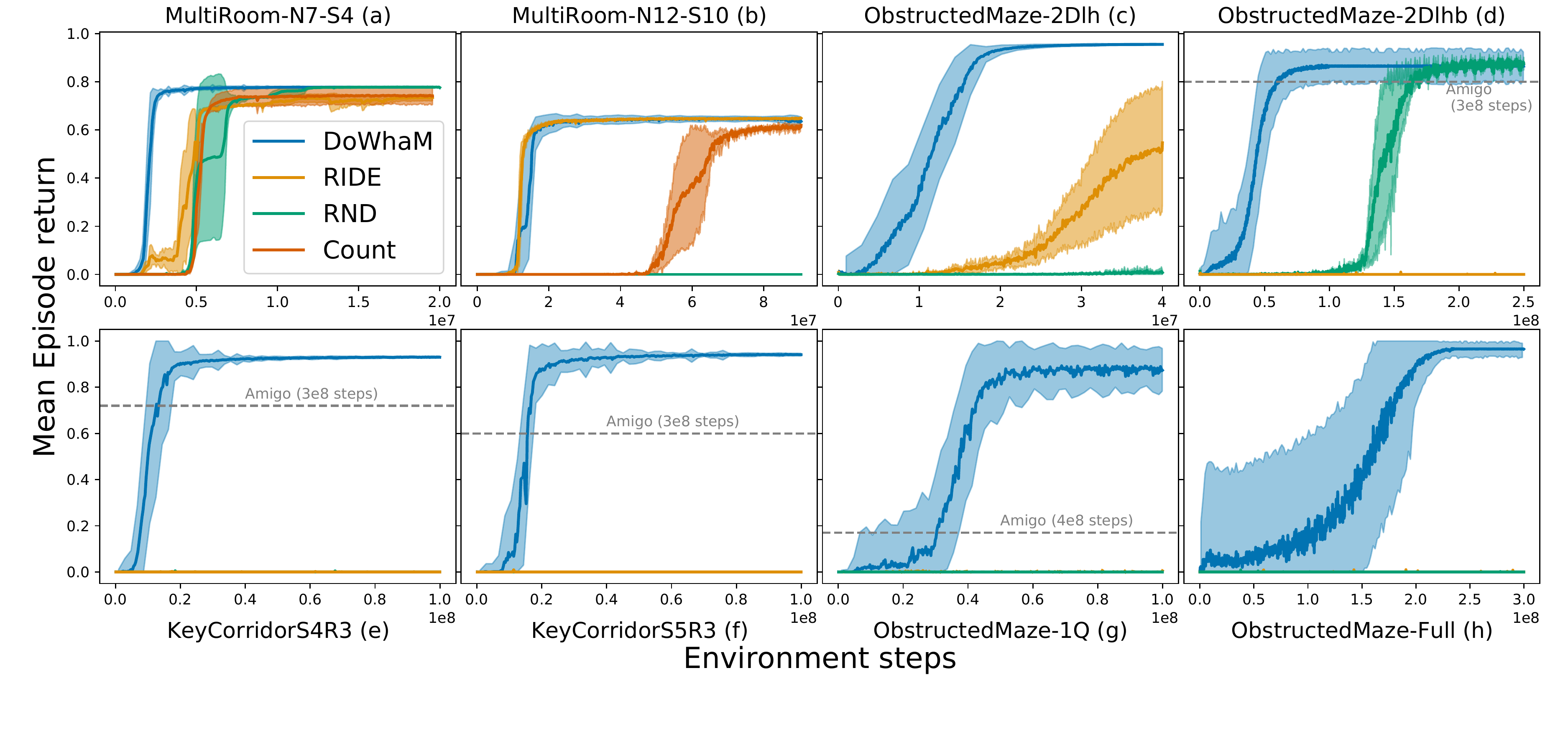}
	\caption{Comparison between intrinsically motivated methods on multiple MiniGrid tasks.}
	\label{fig:compileresults}
\end{figure*}

Each new MiniGrid world contains a combination of rooms that are populated with objects (balls, boxes or keys), and are linked together through (locked/unlocked) doors. Balls and keys can be picked up or dropped and the agent may only carry one of them at a time. Boxes can be opened to discover a hidden colored key. Doors can be unlocked with keys matching their color. The agent is rewarded after reaching the goal tile. At each step, the agent observes a 7x7 representation of its field of view and the item it carries if any. The agent may perform one out of seven actions: move forward, turn right, turn left, pick-up object, drop object, toggle. Noticeably, some actions are ineffective in specific states, e.g. moving forward in front of a wall, picking-up/dropping/toggling objects when none is available. Following~\citep{raileanu2019ride,campero2020learning}, we focus on three hard exploration tasks, which are illustrated in \autoref{fig:compileresults}.


\paragraph{MultiRoom($N$-$S$):} The agent must navigate through a sequence of empty rooms connected by doors of different colors. A map contains $N$ rooms, whose indoor width and height are sampled within $2$ and $S-2$ tiles. MultiRoom maps entail limited interaction as the agent only has to toggle doors and no object manipulation is required. Yet, this bare-bone environment constitutes a good preliminary trial.


\paragraph{KeyCorridor($S$-$R$):} The agent must explore multiple adjacent unlocked rooms to retrieve a key, open the remaining locked room, and collect the green ball. A map contains a large main corridor connected to $2 \times R$ square rooms of fixed indoor dimension $S-2$. Solving a KeyCorridor map requires the agent to perform a specific sequence of interactions, which makes the task more difficult than MultiRoom.


\paragraph{ObstructedMaze:} The agent must explore a grid of rooms that are randomly connected to each others in order to collect a blue ball. Some of the doors are locked and the agent has to either directly collect the keys or toggle boxes to reveal them. Besides, distractor balls are added to block door access. ObstructedMaze can quickly become a hard maze with false leads and complex interactions.


\subsection{Experimental Setting}

\paragraph{Training} We follow the training protocol defined in \citep{raileanu2019ride,campero2020learning}.
We use 3 convolution layers with a kernel size of 3, followed by 2 fully-connected layers of size 1024, and an LSTM of hidden size 1024. Finally, we use two separate fully-connected layers of size 1024 for the actor's and critic's head.
We train our model with the distributed actor-critic algorithm IMPALA~\citep{espeholt2018impala} TorchBeast implementation~\citep{torchbeast2019}. (All parameters in \autoref{subsec:trainingparam}).

\paragraph{Baselines} We here cover the three common families of intrinsically motivated reward mechanisms. 
%
\textcolor{count}{COUNT}~\citep{strehl2008analysis} is a counting method that baits the agent to explore less visited states. In this setting, we use a tabular-count to catalog the state-action pairs.
\textcolor{rnd}{RND}~\citep{burda2018exploration} acts as a states' pseudo-count method. A network is trained to predict randomly projected states and the normalized predicton error is used as intrinsic reward.
%
%
\textcolor{ride}{RIDE}~\citep{raileanu2019ride} is a curiosity-based model that builds upon~\citep{pathak2017curiosity}. It computes the difference between two consecutive states, encouraging the agent to perform actions that lead it to a maximally different states. 
%
\textcolor{gray}{AMIGO}~\citep{campero2020learning} is a hierarchical goal-based method, splitting the agent into two components: an adversarial goal-setter and a goal-condition learner that adversarially creates goals.

\section{Experimental Results}

 \begin{figure}
    \centering
    \includegraphics[width=\linewidth]{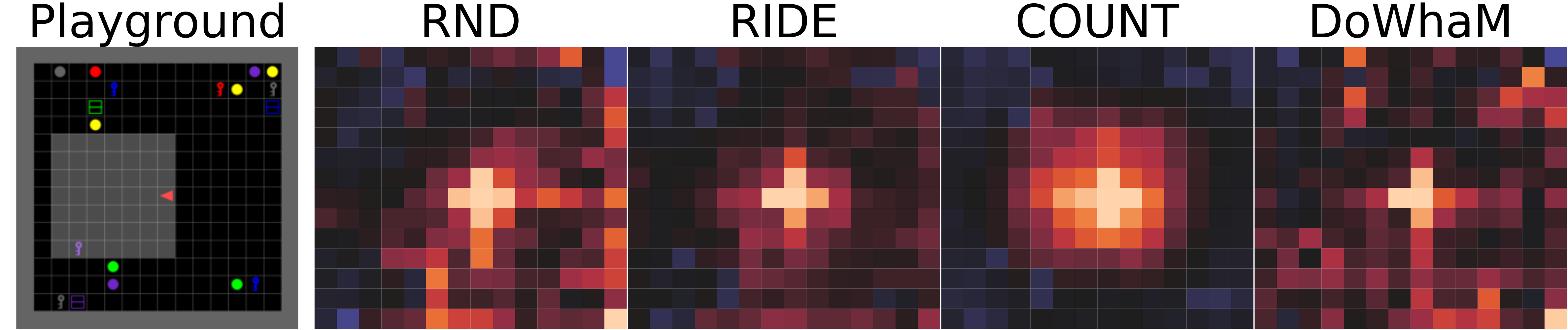}
    \caption{States visitation in Playground environment. \textcolor{bright}{Bright orange} means \textcolor{bright}{more} visits, \textbf{\textcolor{dark}{darker and blue}} means \textbf{less} visits}
    \label{fig:play_statedistrib}        
\end{figure}

\subsection{Base environment}

\autoref{fig:compileresults} displays the results on 8 MiniGrid tasks. Noticeably, \ram outperforms all the baselines in sample complexity, and even solves among the most complex worlds.
In MultiRoom,  we observe that \ram outperforms RIDE, RND, and COUNT in the simple setup (N7S4), and matches RIDE's sample complexity performance on the challenging setup (N12S10). Note \ram does not seem to be penalized by the small amount of possible interactions.
In KeyCorridor and ObstructedMaze, RIDE, RND, and AMIGO learn in the easiest instances but they struggle as the difficulty, i.e. exploration depth, increases as first observed in~\citep{campero2020learning}. On the other hand, \ram consistently solves all the environments, even the challenging ObstructedMaze-Full (Concurrent works achieve similar performances using an adversarial regularization approach~ \citep{flet2021adversarially} or frontier modeling~\cite{zhang2020bebold})

We derive two hypotheses from those results: 
(1) State-count rewards exhaustively explore the state space, reducing the overall exploration coverage
(2) Curiosity-based rewards do not emphasize enough salient interactions and then explore new but irrelevant state-action pairs. 
Although such approaches were successful in many environments, those exploration behaviors may fail as soon as specific interactions must be regularly performed in the exploration process. In the following, we thus try to assess those hypotheses. 



\subsection{Intrinsic exploration behavior}\label{sub:rewardless}

We first conduct a series of experiments without external reward to study what \textit{type of exploration} each bonus creates. In other words, what are the inductive exploration biases that arise from the different intrinsic reward mechanisms. To do so, we rely on two metrics: the state visit (plotted as heatmaps) and the action distribution (plotted as bar plots). 
 
\paragraph{Rewardless Playground} In this spirit, we design a sandbox environment without any specific goal to observe the agent behavior visually, akin to a kindergarten. This environment contains multiple keys, balls, and boxes located in the corners and spawns the agent facing a random direction. \autoref{fig:play_statedistrib} shows the agent state visits for during $10^6$ training timesteps when only using the intrinsic reward signal.

We observe that RND and \ram are both attracted by the objects and explore the space thoroughly, whereas RIDE and COUNT remain close to the center and seldomly reach the objects. This observation backs our results in ObstructedMaze2Dlh, where RND and \ram are the only methods exploring thoughtfully the environment. It also confirms our hypothesis that standard state-based approaches, e.g., COUNT, may not be pushed enough to perform in-depth exploration. Surprisingly, the curiosity-based method RIDE has not been strongly incentivized to interact with remote objects, suggesting that it may suffer from its dependency on the state representation. 
However, these experiments do not explain the performance difference between RND and \ram on the most challenging setups. Thus, looking at the action distribution is necessary.



\begin{figure}[H]
	\centering
	\begin{minipage}{\linewidth}  
		\centering
		\includegraphics[width=0.85\linewidth]{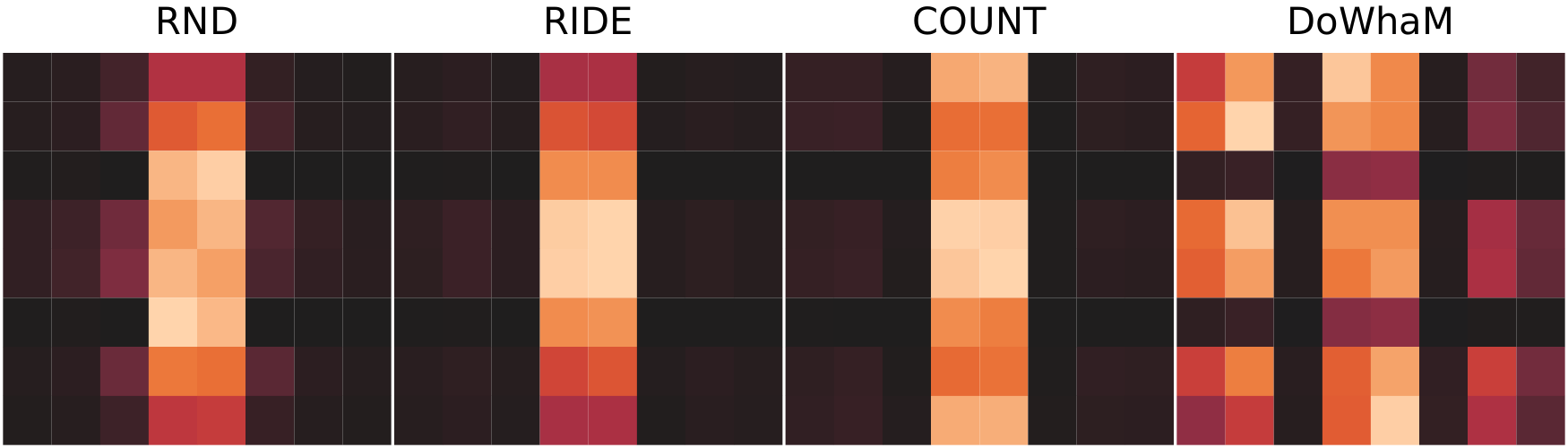}
	\end{minipage} 
	\begin{minipage}{\linewidth}
			\centering
		\includegraphics[width=0.9\textwidth]{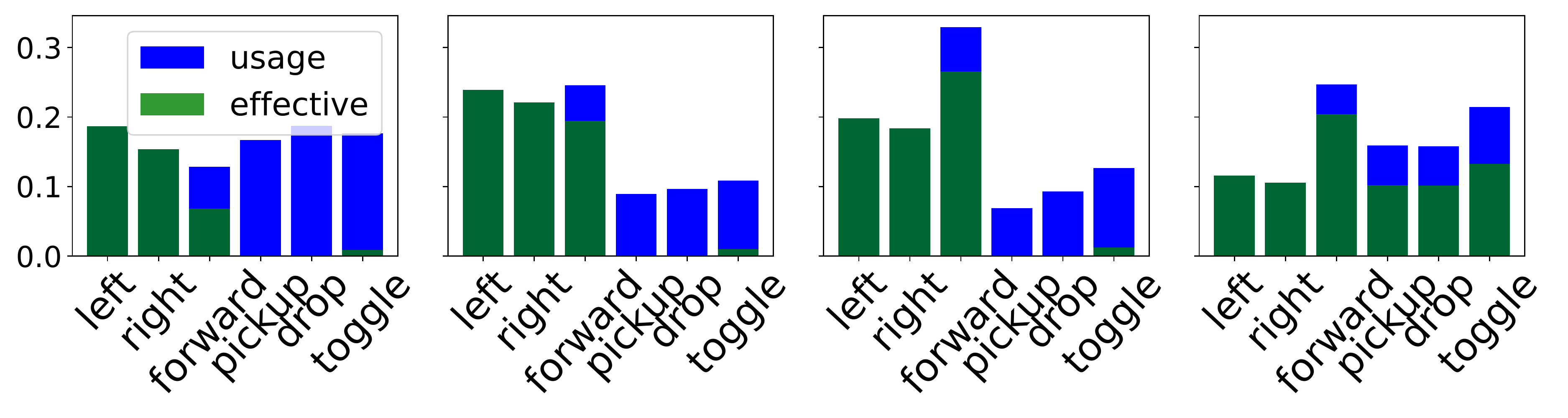}
	\end{minipage}
\caption{State and action distributions in \emph{rewardless} KeyCorridor (S4R3). \textcolor{blue}{$U^{\mathcal{H}}(a)$} and \textcolor{green}{$E^{\mathcal{H}}(a)$} action-count are in blue and green.  Only \ram correctly uses pickup/drop/toggle during exploration.}
\label{fig:kcorr_rewardless}
\end{figure}

\paragraph{Rewardless KeyCorridorS4R3} \label{subsub:kcrewardless}
We then study the behavior that is solely intrinsically motivated in the KeyCorridor environment to better grasp the \ram performance in this setting. Similarly, we trained the agents on KeyCorridorS4R3 for $10^7$ timesteps with only the intrinsic reward signal, and results are displayed in \autoref{fig:kcorr_rewardless}. 

All the baselines -- RIDE, RND, and COUNT -- remain mostly stuck in the central corridor, where \ram explores rooms more uniformly. More impressively, the \ram agent naturally picks the key, enters the locked room, and grabs the ball 7\% of the times without any extrinsic reward. COUNT, RIDE, and RND all have a success ratio below 0.6\%, which may explain why \ram manages to solve this task. Further details can be found in \autoref{fig:forced_path} and \autoref{tab:rewardless_reward}.

We also observe a large discrepancy in the action distribution between the different methods.
First, we observe that RND and \ram action distributions remain approximately uniform while RIDE and COUNT favor moving actions, reducing the opportunity for interactions. Second, and crucially, the impact distributions $E^{\mathcal{H}}(a)$ differs drastically between \ram and other methods. All agents are trying actions such as \textit{pick, toggle} or \textit{drop}, but those actions are rarely changing the agent's state. These actions are not used in the appropriate context, i.e., in front of an object. It means that rewarding state novelty might not be enough to discover effective actions, thus wasting samples. Although \ram and RND had similar state-visitation and action distribution patterns, only \ram correctly apprehend rarely effective actions, and correctly use them to explore its environment.

\subsection{Intrinsic Motivation Pitfalls}\label{sec:intrinsic}

\begin{figure}
    \centering
    \includegraphics[width=\linewidth]{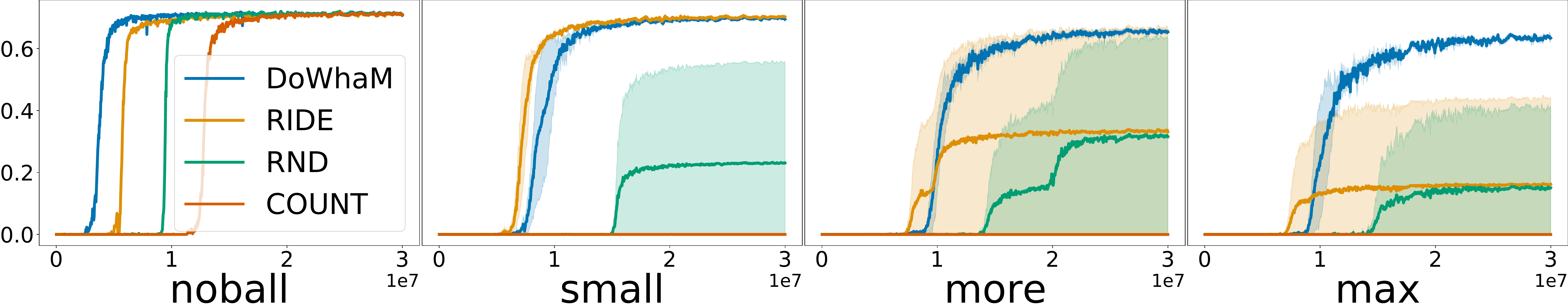}
    \caption{As distractors are added (from left to right), we observe a drop in performance for all methods.}
    \label{fig:ballpit_comparison}
\end{figure}

\paragraph{The Ball Pit Problem}



As \ram biases the state visit distribution towards performing rare actions, it may introduce a poor exploration pattern when facing too many of such states. We refer to this potential issue as the \textit{Ballpit problem}: the agent remains in rooms with plenty of balls to interact with. We created four versions of \textit{Multiroom} (normal, small, more, max), and randomly spawned objects to assess the agent behavior (more details in \autoref{app:env}). 

As the number of objects grows, the performance of all algorithms deteriorates. RND, COUNT are mostly affected by this problem, as the number of states is growing exponentially; thus, counting state occurrence is challenging. RIDE is less affected by the BallPit problem, but most surprisingly, \ram is the only one to reach the exit consistently in the most challenging setup. The $E^{\mathcal{H}}(a)/U^{\mathcal{H}}(a)$ ratio correctly balances the exploration bonus, and does not take over the final extrinsic reward.

\paragraph{The Noisy-TV problem} State-count based agent are attracted to state-action pairs with random noise. In its current definition, \ram is also affected while computing $E^{\mathcal{H}}$. Similar to~\citep{burda2018exploration}, this effect can be circumvented by using an inverse model, and we leave it for future work.

\begin{figure}[h]
	\centering
	\includegraphics[width=\textwidth]{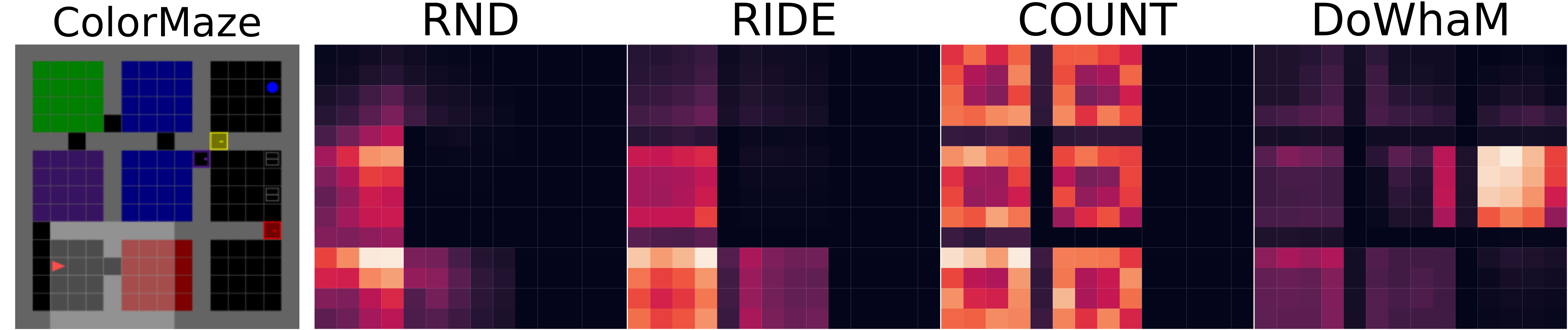}
	\caption{RND, RIDE and COUNT remain within the colored region whereas \ram learns to go straight to the boxes and keys.}
	\label{fig:longobstructed}
\end{figure}

\paragraph{ColorMaze} 
In \autoref{fig:longobstructed}, we design a map with a sequence of open rooms, colored floor changing every episode, two boxes with one hidden key, and a locked door leading to the reward. All baselines remain in the first part of the maze while \ram quickly reaches the objects and solves the task. This experiment highlights again how shifting the emphasis from exhaustive state-visit to relevant state-visit can be beneficial, and change the exploration pattern.



\section{Conclusion}

We introduce \ramlong (\ram), a new action-based intrinsic exploration algorithm. As opposed to count-based and curiosity-driven methods, \ram shifts the emphasis from novel state to state with relevant actions, rewarding actions that are rarely effective in the environment. 
Combined with a simple episodic count, \ram outperforms recent exploration methods on a variety of hard exploratory tasks in a Minigrid environment. This proof of concept illustrates that action-based exploration is a promising approach as it induces surprisingly different exploration patterns. We also pointed out a new category of problems called \textit{BallPit}, which deteriorate performance of many intrinsically motivated reward approaches. 

\section{Acknowledgments}

We would like the thank the anonymous reviewers for useful feedbacks, Bilal Piot and Olivier Teboul for proofreading the manuscript, RIDE's authors for releasing their code and baselines, and finally the stimulating environment of \href{https://team.inria.fr/scool/}{Scool}. Experiments presented in this paper were carried out using the Grid'5000 testbed, supported by a scientific interest group hosted by Inria and including CNRS, RENATER and several Universities as well as other organizations (see \url{https://www.grid5000.fr}).

{
\bibliographystyle{named}
\bibliography{ijcai20}
}


\newpage
\appendix
\section{Appendix}

\subsection{Environments introduced in the paper}\label{app:env}

\textbf{Playground} is designed to visually assess how exploration strategies behave when lots of object are present. Playground is a 14x14 grid, objects are always spawned at the same location but the color changes from episode to episode. An episode lasts 200 steps, no external reward is given and the agent is always spawned in the center, facing a random direction.

\paragraph{Ballpit environment}

The Ballpit environment was designed to assess how each algorithm is affected by the \textit{Ballpit problem}. We build it on top of MultiRoomN4S6, a simple environment solved by all baselines. \textbf{No Ball} is the baseline environment. In \textbf{Small} instances, 1 random object is spawned, \textbf{More} contains 3 random objects and \textbf{Max} rooms are completely filled without blocking the agent path. We emphasize that the agent does not need to interact with anything (with the exception of door) to solve the environment and objects are never in the way of the agent.

\begin{figure}[H]
    \centering
	\includegraphics[width=0.7\linewidth]{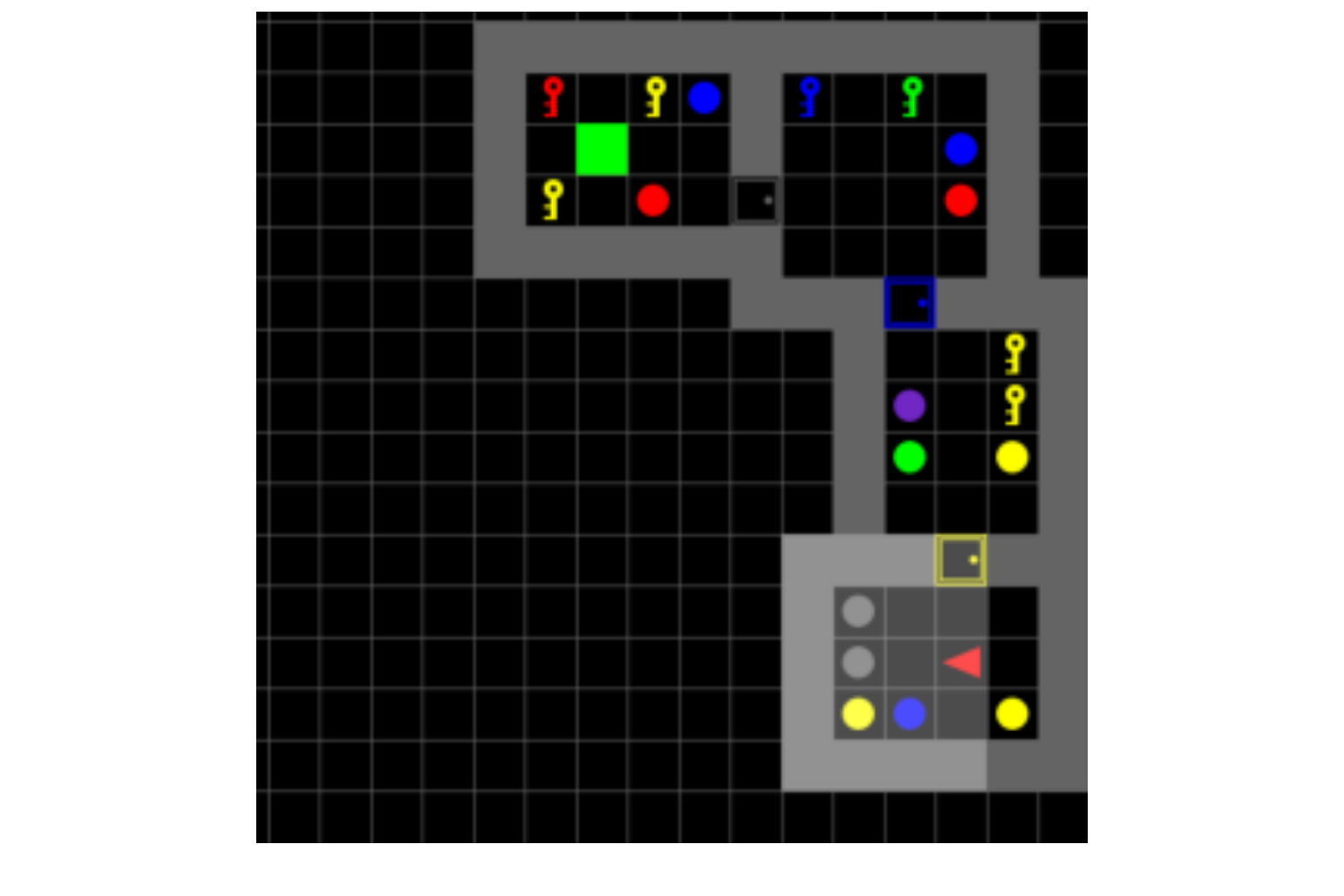}
	\caption{Illustration of BallPit-Max}
	\label{fig:ballpitenv}
\end{figure}

\paragraph{ColorMaze}

ColorMaze is a variation of ObstructedMaze2Dlh. Before accessing the room containing keys within boxes, the agent must cross 4 colored rooms.
A distractor room is placed at the beginning, leading to nothing.
The colors are picked randomly at the beginning of an episode and remains the same until the episode's end. The agent must catch the blue ball within a 576 steps limit to receive the reward. One main difference between MultiRoom and ColorMaze is the absence of doors between colored room, which make this environment harder for \ram as no interaction is possible until the first door in room 4.

\subsection{Decay illustration}\label{app:decay}

\ram reward is a function of the action ratio : $E^{\mathcal{H}}(a)$ the number of times an action impacted the state over $U^{\mathcal{H}}(a)$ the number of usage. The function acts as an exponential decay starting from 1 when the ratio is 0 (the action never impacted the environment so we want to reward it highly when it is actually modifying the state) and giving 0 reward when the ratio is 1 (The action $a$ modifies the state all the time; thus, $a$ is of low interest).

\begin{figure}[H]
    \centering
    \includegraphics[width=0.8\linewidth]{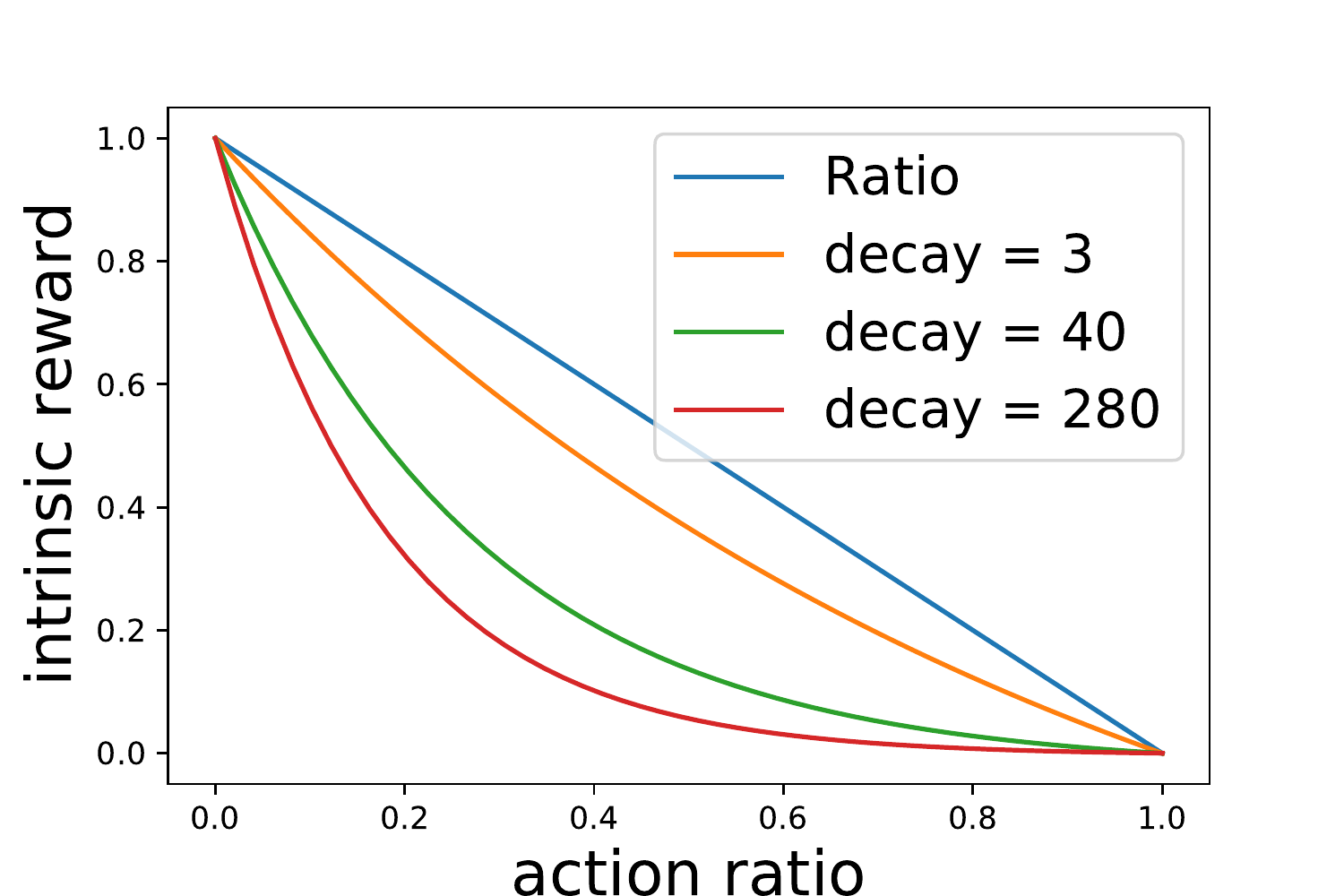}
    \caption{Ratio decay function illustrated for different values of $\eta$}
    \label{fig:decay}
\end{figure}

In \autoref{fig:decay}, we illustrate how the parameter $\eta$ shape the intrinsic reward.

\begin{equation}
	R(s_t, a_t, s_{t+1})= 
		\frac{
		    \eta^{1 - \frac{E^{\mathcal{H}}(a_t)}{U^{\mathcal{H}}(a_t)}} - 1
		    }
		    {\eta - 1}
\end{equation}

\subsection{\ram parameters}\label{subsec:trainingparam}

\begin{table}[H]
\centering
\begin{tabular}{|l|l|}
\hline
\textbf{Parameters}      & \textbf{Value} \\ \hline
Intrinsic reward scaling & 0.05           \\ \hline
Ratio decay  $\eta$      & 40             \\ \hline
Learning rate            & 1e-5           \\ \hline
Optimizer                & RMSprop        \\ \hline
Entropy cost             & 0.001          \\ \hline
batch size               & 32             \\ \hline
LSTM unroll length       & 100            \\ \hline
Number of actors         & 42             \\ \hline
Max norm gradient        & 40             \\ \hline
Baseline cost            & 0.5            \\ \hline
Discount factor          & 0.99           \\ \hline
\end{tabular}
\end{table}

All parameters are kept the same across all environments, except for MultiRoomN12-S10 where the entropy cost is 0.0001.


\begin{figure*}[ht]
	\centering
	\begin{minipage}[t]{\linewidth}	
		\centering
		\begin{minipage}{0.3\textwidth}
			\includegraphics[width=\textwidth]{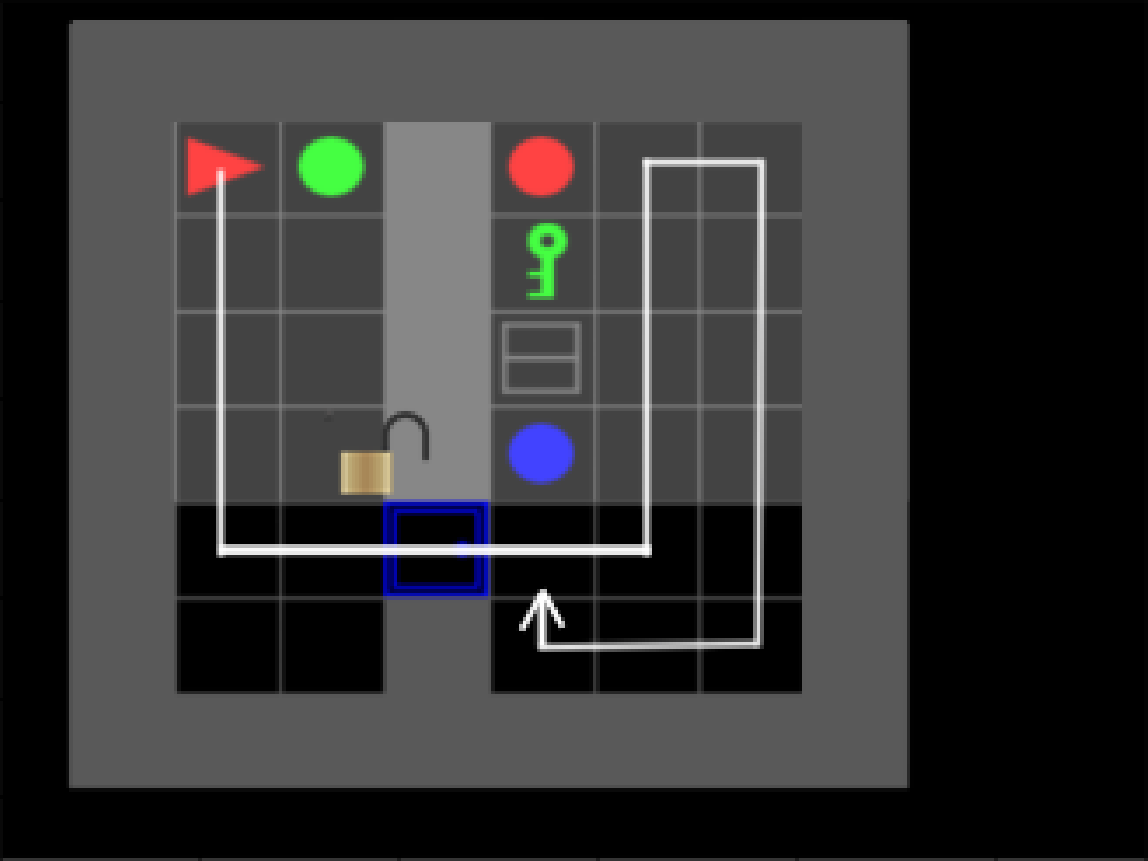}
		\end{minipage}
		\begin{minipage}{0.3\textwidth}
			\includegraphics[width=\textwidth]{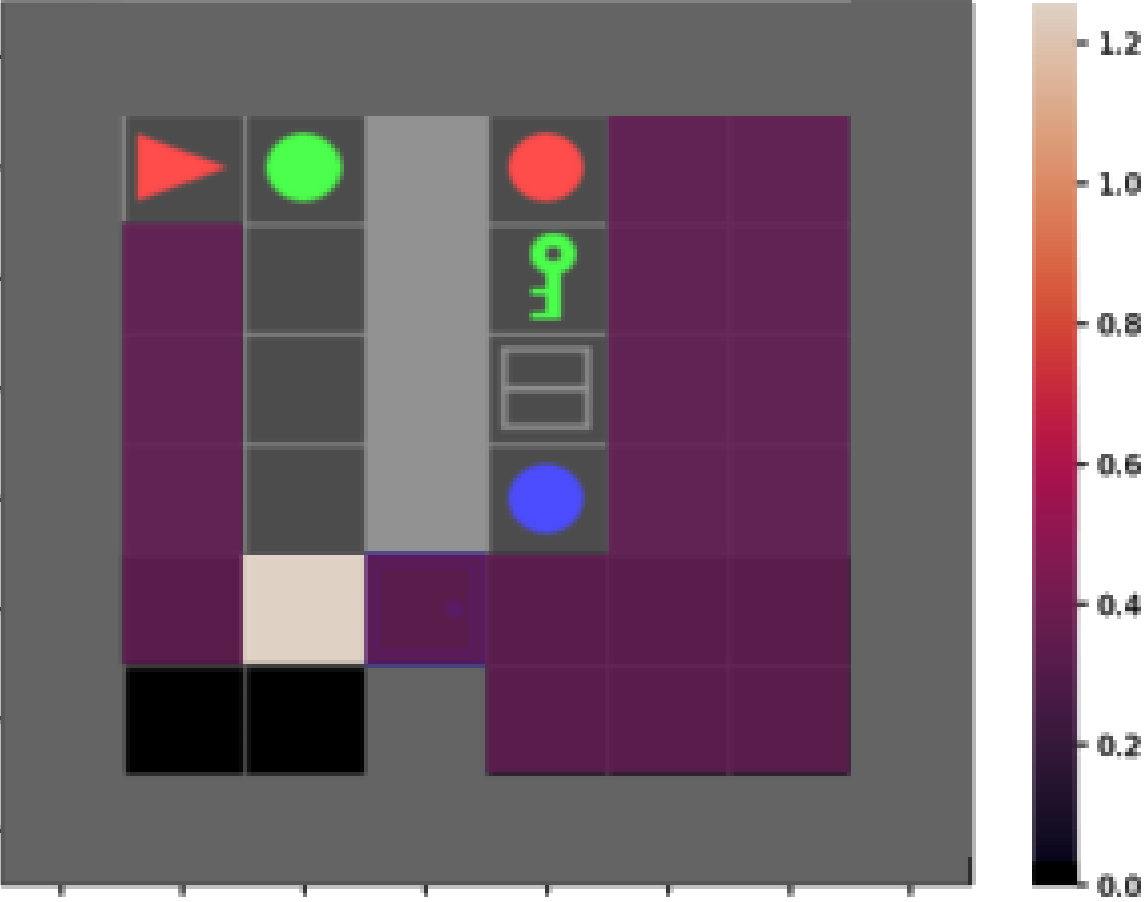}
		\end{minipage}
		\begin{minipage}{0.3\textwidth}
			\includegraphics[width=\textwidth]{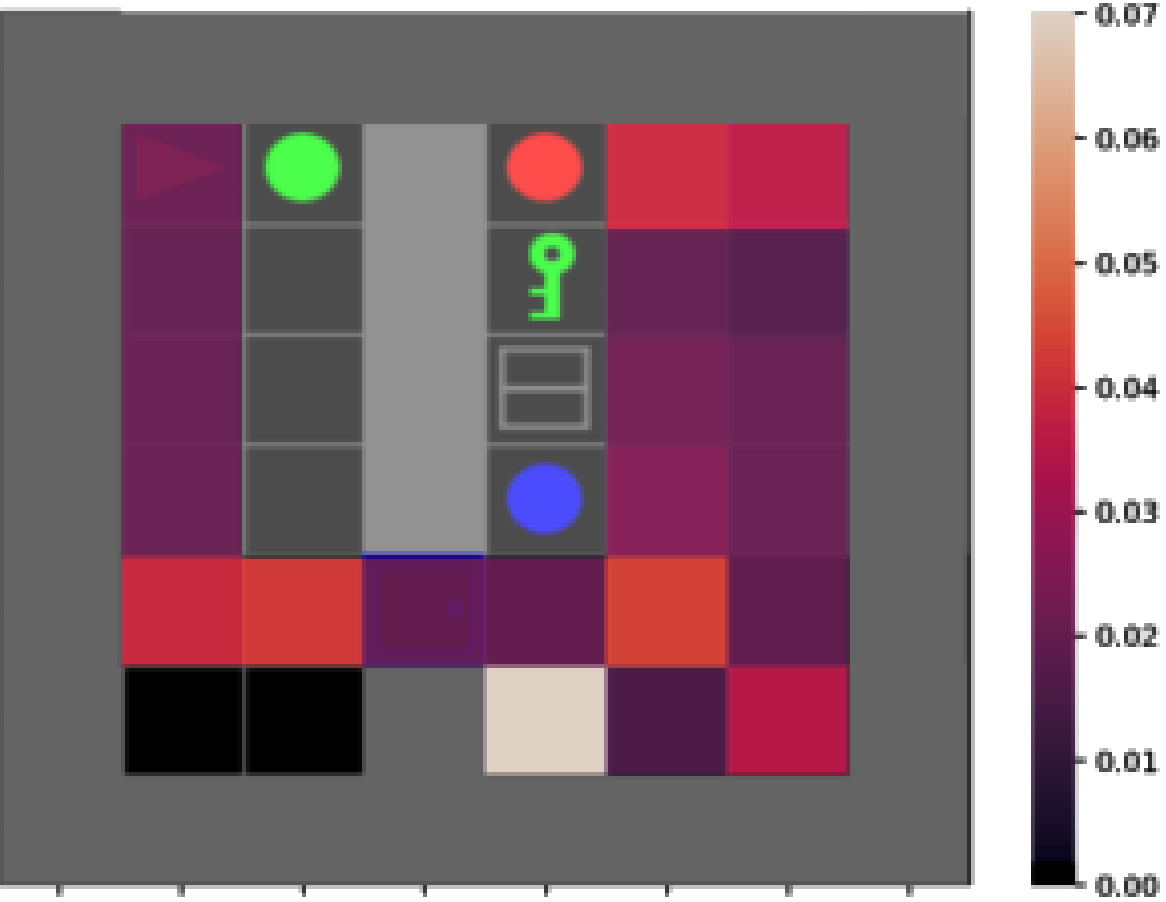}      
		\end{minipage}
	\end{minipage}\\
	\begin{minipage}[t]{\linewidth}
		\centering
		\begin{minipage}{0.3\textwidth}
			\includegraphics[width=\textwidth]{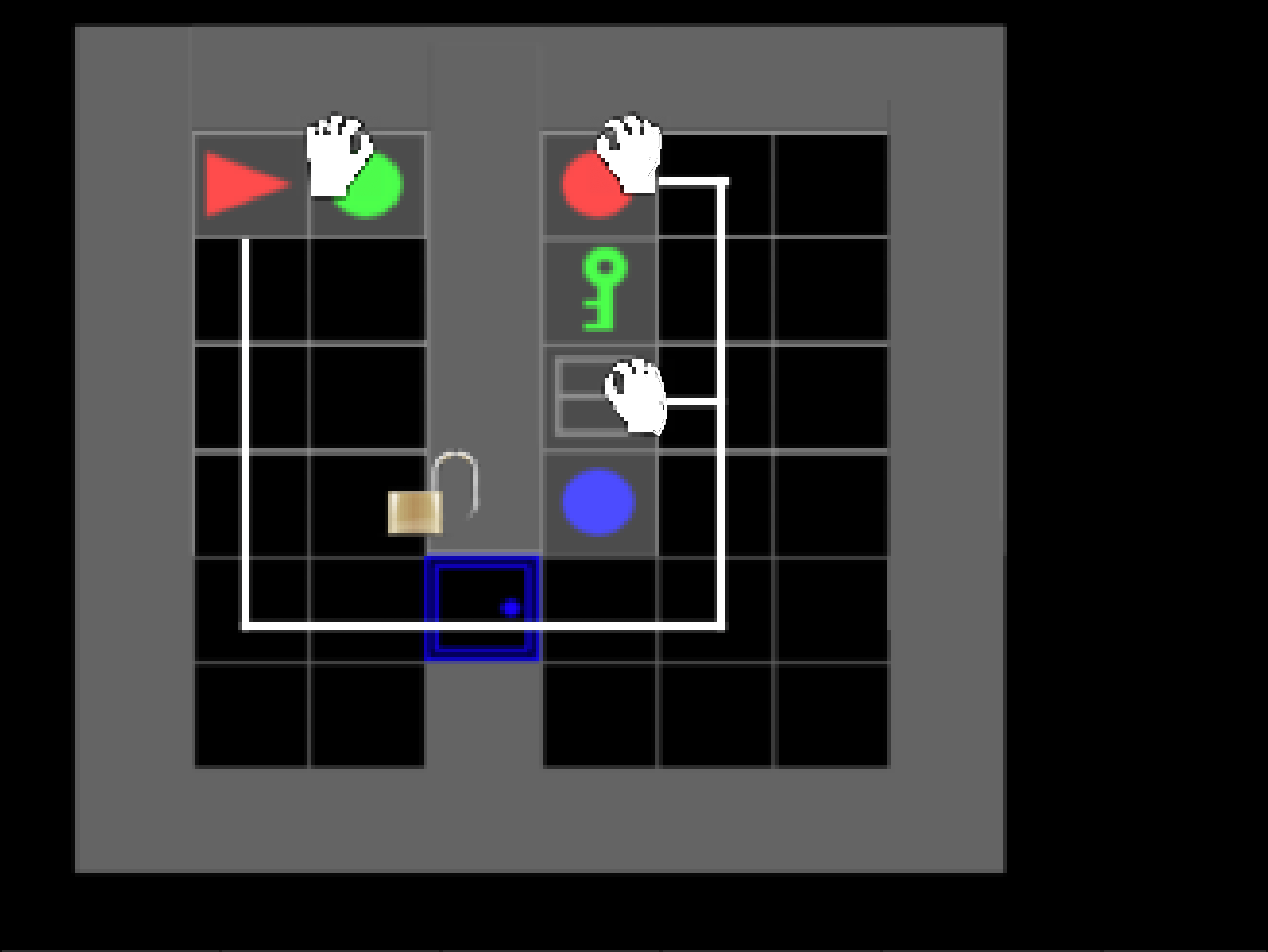}
		\end{minipage}
		\begin{minipage}{0.3\textwidth}
			\includegraphics[width=\textwidth]{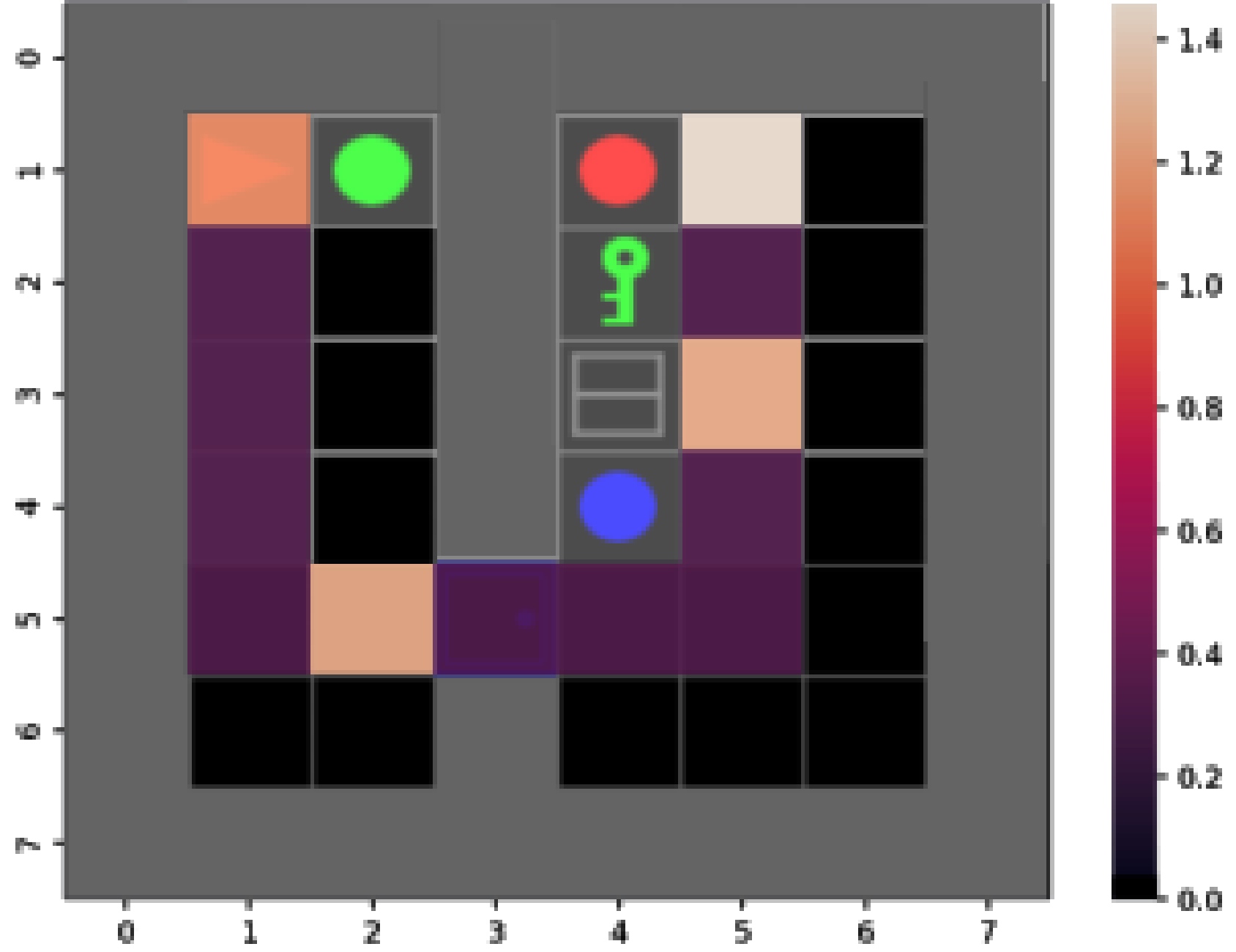}
		\end{minipage}
		\begin{minipage}{0.3\textwidth}
			\includegraphics[width=\textwidth]{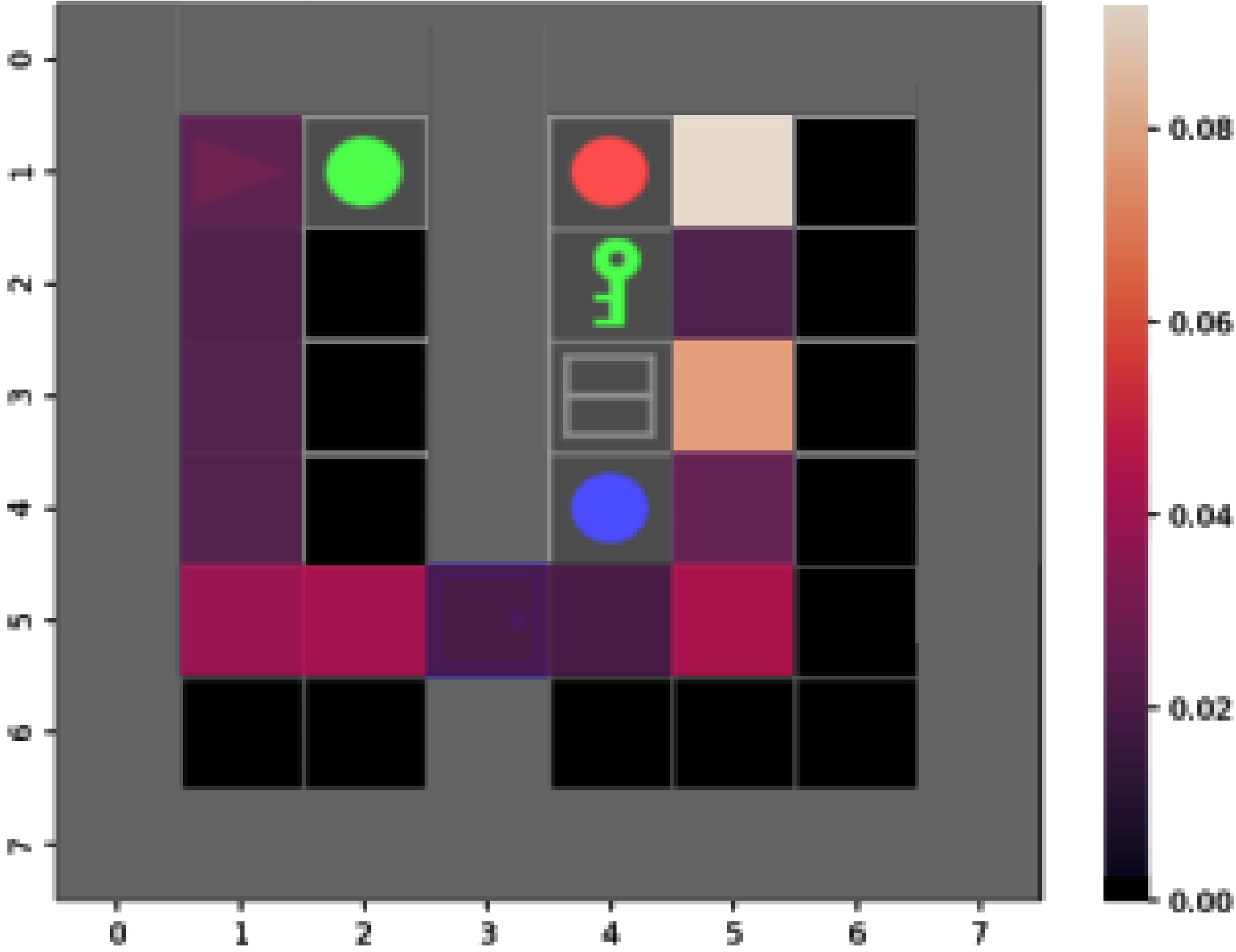}
		\end{minipage}
	\end{minipage}
	\caption{Two scenarios where we force paths and only inspect the intrinsic bonus given. (left) Forced path : Interaction scenario (middle) \ram (left) RIDE. \ram's intrinsic reward is less dense than other state-based methods such as RIDE. It may thus provides intermediate reward milestones that encourage deep exploration, rather than a stream of bonus which may blur salient interaction.}
	\label{fig:forced_path}
\end{figure*}


\begin{table}
	\begin{tabular}{|l|l|l|l|l|}
		\hline
		\textbf{Algorithm}                                                                    & \textbf{RND} & \textbf{COUNT} & \textbf{RIDE} & \textbf{\ram} \\ \hline
		\textbf{\begin{tabular}[c]{@{}l@{}}\% of extrinsic \\ reward collection\end{tabular}} & 0\%          & 0.4\%          & 0.6\%         & \textbf{7\%} \\ \hline
	\end{tabular}
	\caption{Rewardless KeyCorridor : Percentage of times the agent collects extrinsic rewards while only receiving intrinsic rewards (average over 420 episodes)}
	\label{tab:rewardless_reward}
\end{table}

\end{document}

%% file: math_commands.tex
\usepackage{amsmath,amsfonts,bm}

\def\eqref#1{equation~\ref{#1}}

\def\1{\bm{1}}

\DeclareMathAlphabet{\mathsfit}{\encodingdefault}{\sfdefault}{m}{sl}
\SetMathAlphabet{\mathsfit}{bold}{\encodingdefault}{\sfdefault}{bx}{n}

